\documentclass{article}
\usepackage{amssymb}

\usepackage{amsmath}  \usepackage{amsfonts}

\usepackage{enumitem} 
\usepackage{graphicx}
\usepackage[preprint]{corl_2025} 

\usepackage{xcolor}

\definecolor{alexcolor}{RGB}{220, 20, 60}   
\definecolor{emmacolor}{RGB}{0, 128, 255}   
\definecolor{limingcolor}{RGB}{0, 160, 100} 

\title{Object-Centric Representations Improve Policy Generalization in Robot Manipulation}

\author{
  Alexandre Chapin \\
  Ecole Centrale de Lyon, LIRIS \\
  69130, Ecully, France\\
  \texttt{alexandre.chapin@ec-lyon.fr} \\
  \And
  Bruno Machado \\
  Ecole Centrale de Lyon, LIRIS \\
  69130, Ecully, France\\
  \texttt{alexandre.chapin@ec-lyon.fr} \\
  \And
  Emmanuel Dellandrea \\
  Ecole Centrale de Lyon, LIRIS \\
  69130, Ecully, France\\
  \texttt{emmanuel.dellandrea@ec-lyon.fr} \\
  \And
  Liming Chen \\
  Ecole Centrale de Lyon, LIRIS \\
  69130, Ecully, France\\
  \texttt{liming.chen@ec-lyon.fr} \\
}

\begin{document}
\maketitle


\begin{abstract}
Visual representations are central to the learning and generalization capabilities of robotic manipulation policies. While existing methods rely on global or dense features, such representations often entangle task-relevant and irrelevant scene information, limiting robustness under distribution shifts. In this work, we investigate object-centric representations (OCR) as a structured alternative that segments visual input into a finished set of entities, introducing inductive biases that align more naturally with manipulation tasks. We benchmark a range of visual encoders—object-centric, global and dense methods—across a suite of simulated and real-world manipulation tasks ranging from simple to complex, and evaluate their generalization under diverse visual conditions including changes in lighting, texture, and the presence of distractors. Our findings reveal that OCR-based policies outperform dense and global representations in generalization settings, even without task-specific pretraining. These insights suggest that OCR is a promising direction for designing visual systems that generalize effectively in dynamic, real-world robotic environments.
\end{abstract}


\begin{figure}[h]
    \centering
    \includegraphics[width=0.9\textwidth]{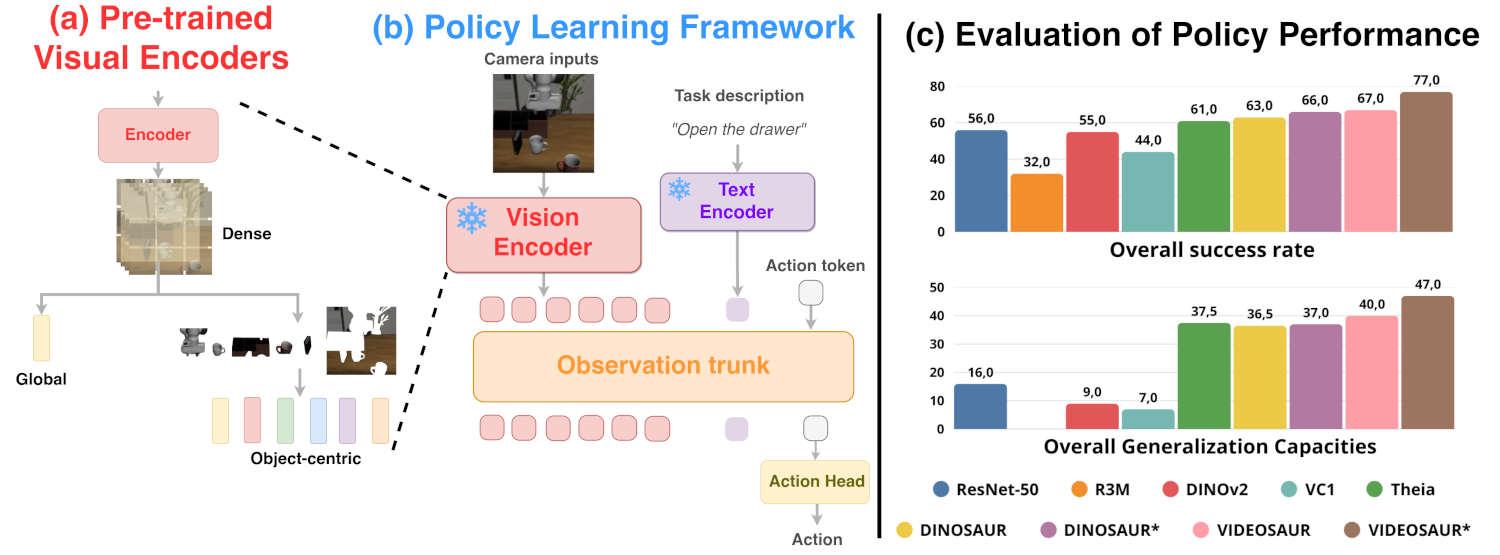}
    \caption{\textbf{Overall architecture.} We use a set of pre-trained visual models with different structures of latent space - global, dense and object-centric - (a) as input to a policy model for robotic manipulation learning (b). We showcase the benefits of Object-Centric Representations (VIDEOSAUR, DINOSAUR) over different visual models on the final performance of policies and the generalization capabilities  in simulation and real-world scenarios (c). DINOSAUR* and VIDEOSAUR* are our versions of the OCR models with the attention module trained using a mixture of robot data. 
    }
    \label{fig:overview}
\end{figure}

\section{Introduction}
Visuomotor policy learning enables robots to map raw visual sensory inputs directly to control actions, allowing them to perceive and interact with their environment. The core objective is to learn policies that are both \textbf{generalizable} and \textbf{sample-efficient}. State-of-the art approaches have so far adopted a common architectural scheme to learn visuomotor policies through imitation learning, dividing the learning process into distinct components: a sensor encoding module, an observation trunk that integrates different modalities, and an action head that predicts the final action \cite{haldar2024bakuefficienttransformermultitask, brohan2023rt1roboticstransformerrealworld, octomodelteam2024octoopensourcegeneralistrobot, kim2024openvlaopensourcevisionlanguageactionmodel}. A key factor in the success of this process is the quality of the visual representation, which must capture task-relevant features while remaining robust to variations in the environment.

As such, recent work has increasingly focused on the role of visual representations in enabling generalizable robot policies. Advances include pre-training on large-scale human egocentric video datasets using time-contrastive and language-aligned objectives \cite{nair2022r3muniversalvisualrepresentation, majumdar2024searchartificialvisualcortex, ma2023vipuniversalvisualreward}, distilling knowledge from diverse vision foundation models into compact, robot-friendly encoders \cite{shang2024theiadistillingdiversevision}, and leveraging powerful self-supervised schemes such as masked autoencoding \cite{radosavovic2022realworldrobotlearningmasked}. Despite these efforts, most prior approaches rely on a shared architectural lineage—typically ResNet or Vision Transformers—and encode visual scenes as  \textbf{global feature vector} (holistic) or  a \textbf{dense patch-based representation} from the encoder's final layers. Such representations often entangle task-relevant and irrelevant information, making them brittle to real-world shifts like lighting changes, novel textures, or cluttered environments \cite{jiang2024robots, parisi2022unsurprisingeffectivenesspretrainedvision, burns2023makespretrainedvisualrepresentations}. 
They are clearly in contrast to human perception, which understands and acts in complex environments by forming symbolic representations of the world around them \cite{lake2016buildingmachineslearnthink}. Rather than processing raw sensory data in isolation, we abstract meaningful entities - "mental symbols" - and reason about them as structured, reusable concepts. This compositional understanding enables us to rapidly generalize from prior experiences to new, unseen situations. At the heart of this capability is our ability to disentangle, bind, and manipulate information flexibly - a trait that remains elusive in today's artificial systems \cite{greff2020bindingproblemartificialneural}. 

There is a growing body of evidence that highlights the need to rethink visual scene representations in robotics, emphasizing structure, abstraction, and task-relevance to bridge the gap between perception and control \cite{kroemer2020reviewrobotlearningmanipulation, bengio2014representationlearningreviewnew}.
A promising approach toward such a capability is \textbf{object-centric representations} (OCRs) which have emerged recently in computer vision \cite{locatello2020objectcentriclearningslotattention, seitzer2023bridginggaprealworldobjectcentric}. These methods restructure images by segmenting them into a set of entities - objects - thereby embedding an inductive bias toward symbolic, compositional reasoning. OCRs introduce structure into the visual pipeline, allowing models to parse scenes into meaningful parts and reason over them explicitly. Notably, this approach mirrors how humans perceive and interact with the world - not as a flat array of pixels, but as a collection of distinct, interactive entities that can be tracked, manipulated, and abstracted into concepts. In this sense, object-centric perception may offer a pathway toward bridging the gap between low-level visual input and high-level symbolic reasoning for robotic manipulation.

Although the potential of OCRs in robotics has been increasingly recognized, much of the existing work has focused on their utility for image decomposition or scene reconstruction. A few recent efforts have explored their role in control, particularly in reinforcement learning tasks \cite{yoon2023investigationpretrainingobjectcentricrepresentations, heravi2023visuomotorcontrolmultiobjectscenes, haramati2024entitycentricreinforcementlearningobject}. However, these studies are often constrained to highly simplified, toy settings \cite{watters2019cobradataefficientmodelbasedrl, kipf2022conditionalobjectcentriclearningvideo, zhang2022objectcentricvideorepresentationbeneficial}, leaving open some critical questions:

\begin{enumerate}[label=Q\arabic*:]
    \item Can OCR models better enable  robotic manipulation policy learning over other visual representations ?
    \item Can OCRs enhance policy generalization capabilities in the presence of  distractors, new textures, and lighting conditions?
\end{enumerate}


To answer these questions, we  make the following contributions:
\begin{itemize}
\item We develop and open-source a unified framework to evaluate different types of visual representations for robotic manipulation.
\item We benchmark 7 distinct visual encoders, including global, dense and object-centric representations, across two simulated environments and a novel, easily replicable real-world set of tasks.
\item Unlike prior works, we show that slot-based object representations enable better robot manipulation policy learning and improve generalization in robot control, especially under realistic distractors and domain shifts, without requiring task-specific tuning.
\end{itemize}


\section{Related works}
\paragraph{Pretrained vision based models for robot learning} In recent years, the computer vision community has developed large-scale pretraining strategies for visual representation learning, yielding models such as MoCo \cite{chen2020improvedbaselinesmomentumcontrastive}, DINO \cite{caron2021emergingpropertiesselfsupervisedvision}, DINOv2 \cite{oquab2024dinov2learningrobustvisual}, and CLIP \cite{radford2021learningtransferablevisualmodels}. These models, originally trained for classification and image-text alignment, have been shown to be surprisingly effective for downstream visuomotor policy learning in robotics \cite{parisi2022unsurprisingeffectivenesspretrainedvision}, even without access to domain-specific robot data.

To better align visual representations with manipulation tasks, follow-up work explored pretraining on more relevant datasets and learning objectives in terms of manipulation. R3M \cite{nair2022r3muniversalvisualrepresentation} learns from large-scale human egocentric video (Ego4D \cite{grauman2022ego4dworld3000hours}) using a combination of time-contrastive learning, video-language alignment, and sparsity regularization, enabling efficient imitation learning in both simulation and the real world. VC-1 \cite{majumdar2024searchartificialvisualcortex} combines masked autoencoding with mixed-domain training (ImageNet and robot video), while Theia \cite{shang2024theiadistillingdiversevision} distills multiple vision foundation models (e.g., CLIP, SAM, DINOv2) into a compact transformer backbone optimized for robotic policy learning.

Recent work has questioned whether robot or manipulation-specific datasets always lead to better pretraining. In a comprehensive study, \textit{Dasari et al.} \cite{dasari2023unbiasedlookdatasetsvisuomotor} found that curated general-purpose datasets such as ImageNet, Kinetics, and 100 Days of Hands often outperform Ego4D and RoboNet when used for visuo-motor pretraining—even under identical architectures and objectives. The study showed that image distribution quality and diversity mattered more than dataset size or domain match, and that popular assumptions about dataset alignment (e.g., egocentric manipualtion data = better for robotics) may not always hold in practice. Similarly, \cite{burns2023makespretrainedvisualrepresentations} showed that vision transformers with strong emergent object segmentation properties generalize better under distribution shifts, such as changes in lighting, textures, and distractors.

These findings motivate a rethinking of visual representations for robot learning—moving away from flat global or dense feature maps toward more structured alternatives. In particular, object-centric representations, which segment visual inputs into discrete entities, promise to capture task-relevant structure more robustly and compositionally. Our work builds on this direction, introducing object-centric biases into visual pretraining and evaluating their impact on generalization and robustness in visuomotor policy learning.

\paragraph{Object-centric learning} Object-Centric Representations (OCR) aims to decompose images into structured representation composed of multiple vectors, commonly referred to as \textit{slots}, each corresponding to an extracted \textit{entity}. There has been recently a growing interest in these methods due to their potential benefits in various domains, including autonomous driving \cite{hamdan2024carformerselfdrivinglearnedobjectcentric}, robotics \cite{heravi2023visuomotorcontrolmultiobjectscenes, mosbach2025soldslotobjectcentriclatent},  and explainability \cite{wang2024explainableimagerecognitionenhanced}. Early OCR research primarily focused on structured representation learning through the lens of generative modeling \cite{kabra2021simoneviewinvarianttemporallyabstractedobject}. Subsequent works adopted encoder-decoder architectures to obtain structured and disentangled latent spaces \cite{burgess2019monetunsupervisedscenedecomposition, kabra2021simoneviewinvarianttemporallyabstractedobject, watters2019cobradataefficientmodelbasedrl}. The most significant method in this field is Slot-Attention \cite{locatello2020objectcentriclearningslotattention}, recognized for its simplicity and efficiency. Further advancements have incorporated more powerful decoding mechanisms, such as diffusion models \cite{jiang2023objectcentricslotdiffusion, wu2023slotdiffusionobjectcentricgenerativemodeling}, transformer decoders \cite{singh2022illiteratedallelearnscompose} or alse pre-trained backbones such as \cite{caron2021emergingpropertiesselfsupervisedvision} to improve OCR performance in real-world scenarios \cite{seitzer2023bridginggaprealworldobjectcentric}. More recently, recurrent neural networks and transformer architectures have been integrated into OCR methods \cite{elsayed2022saviendtoendobjectcentriclearning, kipf2022conditionalobjectcentriclearningvideo, singh2022simpleunsupervisedobjectcentriclearning, zadaianchuk2023objectcentriclearningrealworldvideos} to make them usable for videos.
While OCR naturally segments input images into meaningful components, potentially aiding generalization in imitation learning \cite{jiang2024robots, burns2023makespretrainedvisualrepresentations}, most studies have focused on image reconstruction or semantic segmentation rather than control tasks. Only a few exceptions have explored OCR in simple control environments \cite{haramati2024entitycentricreinforcementlearningobject, heravi2023visuomotorcontrolmultiobjectscenes, yoon2023investigationpretrainingobjectcentricrepresentations}. 

Our goal in this paper is to evaluate whether OCR methods, with their inherent segmentation capabilities, can facilitate visuomotor policy learning and enhance the generalization of robotic manipulation models. 

\section{Method}
\label{sec:method}
The overview of our framework (Figure \ref{fig:overview}) consists of a sensor encoding module (Vision encoder), an observation trunk that integrates different modalities (e.g., text for instructions) and an action head that predicts the final action. We start by introducing an object-centric representation for videos and its variants for visual encoding, followed by policy training.

\paragraph{Object-centric representation for videos}
Object-centric methods have gained traction in computer vision for their ability to generalize across scenes \cite{didolkar2024zeroshotobjectcentricrepresentationlearning} by modeling object-level dynamics and interactions. Consider an input image $O$, the goal is to produce a set of object representations $S = \{s_1, \ldots, s_K\}$. The image is first encoded using a vision backbone into a set of dense features $F = \{f_1, \ldots, f_N\}$ (with $N >> K$), and object-specific representation $S$ -or slots- are extracted using Slot Attention \cite{locatello2020objectcentriclearningslotattention}. Slot Attention is a differentiable module that performs iterative attention with competition, allowing distinct slots to specialize in representing different parts of the input scene. Conceptually, Slot-Attention simply use a cross-attention module \cite{vaswani2023attentionneed}, with a renormalization over queries by Eq. (1). Slots are then obtained by a weighted sum of the values using the attention weights through Eq. (2):

\begin{minipage}{0.70\linewidth}
\begin{equation}
\mathbf{A} = \mathrm{softmax}\left(\frac{\mathbf{Q}\mathbf{K}^T}{\sqrt{D}}\right), \mathbf{Q} \in \mathbb{R}^{N \times D}, \mathbf{K} \in \mathbb{R}^{K \times D}
\end{equation}
\end{minipage}
\hfill
\begin{minipage}{0.20\linewidth}
\begin{equation}
S^{(i+1)} = \mathbf{A}\mathbf{V}
\end{equation}
\end{minipage}


The query $\mathbf{Q}$ in Eq.(1) is the projected set of Slots $S^{(i)}$ at iteration $i$, key $\mathbf{K}$ and values $\mathbf{V}$ are projected dense features $F$.
This process is repeated iteratively in order to obtain the final slots. The original Slot-Attention model used a simple Convolutionnal neural network as visual backbone followed by a simple Deconvolutionnal decoder in order to reconstruct the input image from the slots as learning signal. This method is limited in real-world scenarios with complex images. \textbf{DINOSAUR} \cite{seitzer2023bridginggaprealworldobjectcentric} introduces the use of a frozen \textbf{DINO} \cite{caron2021emergingpropertiesselfsupervisedvision} as visual backbone. The model also reconstruct the extracted features instead of the input image.
\textbf{VIDEOSAUR} \cite{zadaianchuk2023objectcentriclearningrealworldvideos} develop DINOSAUR to videos with two main changes: adding a transformer \textit{predictor} model in order to init the slots of timestep $t$ from the slots predicted from timestep $t-1$ and adding a temporal consistency loss using feature similarity of the updated \textbf{DINOv2} \cite{oquab2024dinov2learningrobustvisual} along two consecutive timesteps


\paragraph{Policy training}
Our primary objective is to assess the effectiveness of object-centric representations when used as inputs for robot policy learning, more specifically through the imitation learning framework. Given a set of expert demonstrations $\mathcal{D} = \{\tau_1, \ldots, \tau_n\}$ with $\tau_i = [(o_0, a_0), \ldots, (o_T, a_T)]$, the policy $\pi$ needs to learn a mapping between the input observation $o_t$ (e.g.,visual inputs) to the next action $a_t$. To ensure a fair comparison, we adopt policy architectures that can process global, dense and slot-based inputs within a unified framework. Moreover, in line with prior work \cite{nair2022r3muniversalvisualrepresentation, burns2023makespretrainedvisualrepresentations, jiang2024robots}, we keep the \textbf{pre-trained visual models frozen} during the policy training process.

For simulated environments, we use the BAKU architecture \cite{haldar2024bakuefficienttransformermultitask}, which consists of an encoder, an observation trunk, and a policy head. The observation trunk is a transformer-based model that encodes a sequence of $T$ past observations. Each observation includes visual features, language embeddings, and proprioceptive states, interleaved with learnable action tokens. The policy head, a deterministic Multi-Layer Perceptron, uses the final action token to produce the next action prediction.

For real-world experiments, we build on Action Chunking Transformers (ACT) \cite{zhao2023learningfinegrainedbimanualmanipulation} as implemented in the LeRobot library\footnote{https://github.com/huggingface/lerobot}. The original model was using dense input representations. We then modified the input modality of ACT to also incorporate either a single global visual feature vector or a set of object-centric slot vectors. To support multi-task learning, we concatenate a task-specific language embedding derived from the ModernBERT model \cite{warner2024smarterbetterfasterlonger} with the visual and proprioceptive inputs. This joint representation is then processed by a transformer encoder-decoder to generate action chunks. Detailed architectural configurations are provided in Appendix \ref{annex:baseline_detail}.


\section{Benchmarks and Experimental setup}
\label{sec:result}

\paragraph{Environments and tasks}



\begin{figure}[h]
    \centering
    \includegraphics[width=\textwidth]{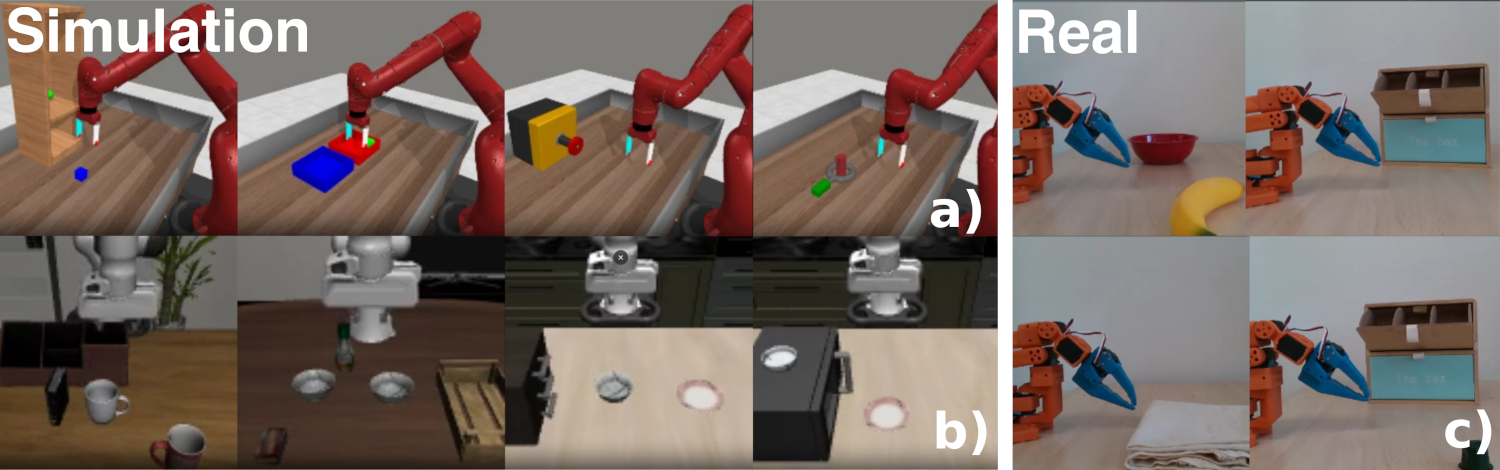}
    \caption{\textbf{Overview environments.} We evaluate the different visual models in two simulation (a: MetaWorld, b: LIBERO) and one real-world environments (c).}
    \label{fig:tasks}
\end{figure}

To comprehensively evaluate visual representations in robot manipulation, we selected three complementary environments—two in simulation and one in the real world—chosen for their diversity in task complexity, embodiment, and visual structure, as shown in Figure~\ref{fig:tasks}.

In simulation, we use MetaWorld \cite{yu2021metaworldbenchmarkevaluationmultitask}, a well-established benchmark comprising tabletop manipulation tasks performed with a Sawyer robotic arm. MetaWorld offers a controlled setting with standardized tasks and supports structured generalization testing, making it a suitable baseline to evaluate representation performance in simple, single-object scenarios.

To challenge the scalability of object-centric representations, we also include LIBERO-90 \cite{liu2023libero}, a recent benchmark featuring multi-object, visually complex scenes across diverse domains such as kitchens, offices, and living rooms. LIBERO is particularly well-suited for our study, as it emphasizes multi-object reasoning and generalization to novel combinations of objects—conditions under which object-centric models are expected to outperform traditional alternatives. 

For real-world evaluation, we introduce a suite of five tabletop manipulation tasks implemented using the LeRobot library and a SO-100 low-cost robotic arm. These tasks were designed to reflect common household activities, such as picking, placing, folding, and drawer manipulation, and involve varied object properties including deformable, articulated, and rigid elements. The setup enables us to test the practical viability of object-centric models under realistic noise and embodiment constraints. All environments and tasks are described more in detail in Annex~\ref{annex:env_detail}.

\paragraph{Pre-trained visual models} Based on recent state-of-the-art evaluations \cite{burns2023makespretrainedvisualrepresentations, majumdar2024searchartificialvisualcortex, jiang2024robots}, we selected 7 pre-trained visual models for comparison that we depict in Table \ref{tab:models}. These models span a range of backbone architectures - including ResNet variants \cite{he2015deepresiduallearningimage} and Vision Transformer-based models \cite{dosovitskiy2021imageworth16x16words} - as well as diverse training objectives such as supervised learning, self-supervised learning, contrastive learning and distillation.  All selected models have demonstrated state-of-the-art performance within their respective categories in benchmark evaluations. We also aimed to capture a broad spectrum of feature representations to support a comprehensive comparison such as the classical global and dense representations but also object-centric representations. Further detail on each model is provided in Appendix \ref{annex:mod_detail}.

\paragraph{Robotic pre-training}
Although object-centric video models show promise in structured scene decomposition, they are typically trained on internet-scale \textit{in-the-wild} video datasets, which depict data diversity but are often misaligned with the distribution of robotic manipulation environments. To bridge this gap and further increase training data diversity in line with the findings in  \cite{dasari2023unbiasedlookdatasetsvisuomotor}, we introduce a dedicated pretraining stage on robotic video datasets.

We compile and preprocess a collection of real-world robotic datasets spanning a wide variety of manipulation skills, environments, and embodiments. In total, our training set includes over 188,000 trajectories, drawn from three major sources: BridgeData V2 \cite{walke2024bridgedatav2datasetrobot}, a diverse set of demonstrations across household tasks using the WidowX-250 arm; Fractal \cite{brohan2023rt1roboticstransformerrealworld}, a large-scale dataset collected with a fleet of Everyday Robots across hundreds of kitchen manipulation tasks; and DROID \cite{khazatsky2024droidlargescaleinthewildrobot}, which contains unconstrained robot interactions across multiple labs and setups. The combined dataset offers rich visual and physical diversity, including varied viewpoints, object types, and lighting conditions. 

To adapt the object-centric model to the robotics domain, we train its slot attention module using a self-supervised reconstruction loss applied to DINO feature maps extracted from temporal video slices. This approach enables the model to learn structured representations grounded in robot manipulation dynamics. In addition, we systematically evaluate how dataset composition impacts downstream performance by comparing models trained on single-source datasets to those trained on a balanced mixture of all three sources (Robot Mixt.). Further details on each dataset and the effects of these training regimes on generalization and policy learning are presented in Annex~\ref{annex:data}.


\begin{table}[h!]
\centering
\caption{\textbf{Models overview.} Comparison of the data and sizes of the different pre-trained visual models . ViT: Vision Transformer, OC: Object-Centric layers, G: Global, D: Dense. DINOSAUR* and VIDEOSAUR* are our versions of the models with the attention module trained on robotic data.}
\begin{tabular}{|c||c|c|c|c|}
\hline
\textbf{Model} & \textbf{Backbone}  & \textbf{Pre-training Dataset} & \textbf{\# of params.} & \textbf{Features}\\
\hline
ResNet-50 \cite{he2015deepresiduallearningimage} & ResNet-50 &ImageNet \cite{russakovsky2015imagenetlargescalevisual}  & 25.6M & D \\
R3M \cite{nair2022r3muniversalvisualrepresentation}& ResNet-50  & Ego4D \cite{grauman2022ego4dworld3000hours} & 25.6M  &  D\\
DINOv2 \cite{oquab2024dinov2learningrobustvisual} & ViT & LVD-142M \cite{oquab2024dinov2learningrobustvisual} & 86M  & G \& D\\
VC-1 \cite{majumdar2024searchartificialvisualcortex}& ViT & Ego4D \cite{grauman2022ego4dworld3000hours} and ImageNet \cite{russakovsky2015imagenetlargescalevisual}  & 86M & G \\
Theia \cite{shang2024theiadistillingdiversevision} & ViT & ImageNet \cite{russakovsky2015imagenetlargescalevisual}  & 140M & D\\
DINOSAUR \cite{seitzer2023bridginggaprealworldobjectcentric} & ViT + OC & COCO \cite{lin2015microsoftcococommonobjects} & 88M  & Slot \\
DINOSAUR* \cite{seitzer2023bridginggaprealworldobjectcentric} & ViT + OC & Robot Mixt. & 88M  & Slot \\
VIDEOSAUR \cite{zadaianchuk2023objectcentriclearningrealworldvideos} & ViT + OC & Youtube-VOS \cite{xu2018youtubevoslargescalevideoobject} & 90M & Slot \\
VIDEOSAUR* \cite{zadaianchuk2023objectcentriclearningrealworldvideos} & ViT + OC & Robot Mixt. & 90M & Slot \\
\hline
\end{tabular}
\label{tab:models}
\end{table}


\section{Results}
\label{sec:result}

We evaluate the role of visual representations in learning and generalizing robotic manipulation policies, focusing specifically on object-centric representations (OCRs) and how they compare to dense and global alternatives. Our results aim to answer the following questions introduced in Section~1: \
\textbf{Q1:} Can OCRs better support robot policy learning than other visual representations? \
\textbf{Q2:} Can OCRs enhance policy generalization under viusal distribution shifts such as distractors, lighting changes, or new textures?

We assess models in both simulation and real-world settings. In simulation, we report mean success rates over three random seeds with 50 rollouts per task; in the real world using LeRobot, we conduct 10 rollouts per task. All policies are trained using frozen visual encoders.

\begin{figure}[h]
    \centering
    \includegraphics[width=0.9\textwidth]{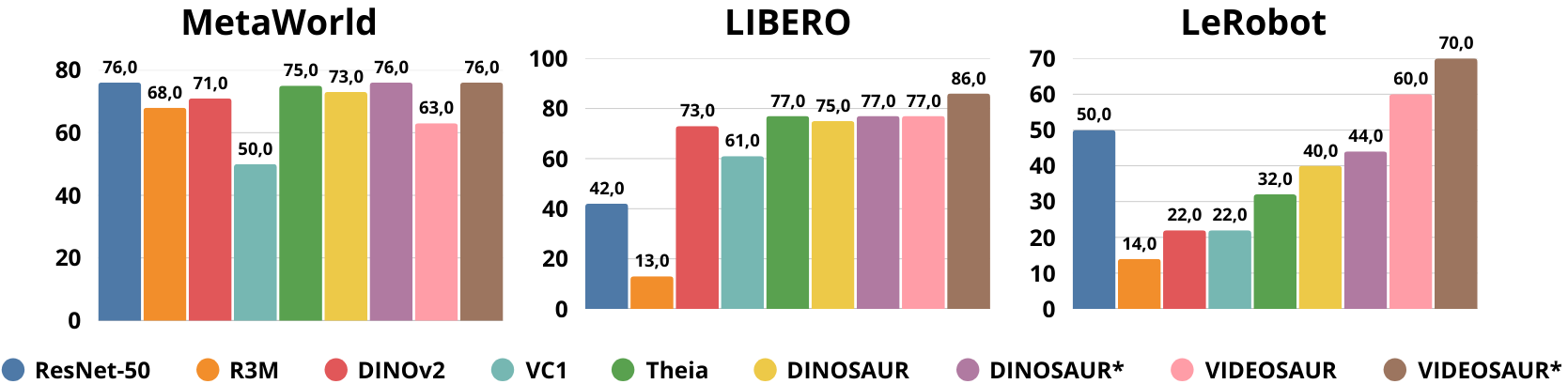}
    \caption{\textbf{Overall success rate.} Mean success rate over all tasks for each visual model on the three environments, e.g., MetaWorld (left), LIBERO (middle) and Real using LeRobot (right). DINOSAUR* and VIDEOSAUR* have  been pretrained over robot data mixture.}
    \label{fig:perf}
\end{figure}

\subsection{Q1: Do OCRs improve manipulation policy learning?}

Figure~\ref{fig:perf} summarizes the average success rate across MetaWorld, LIBERO, and our real-world benchmark. Object-centric models---especially \textbf{VIDEOSAUR*}---consistently achieve the highest overall performance, outperforming both dense (e.g., DINOv2, Theia) and global (e.g., ResNet-50, VC-1) baselines.

In \textbf{MetaWorld}, all models except VC-1 perform above 60\%. The low VC-1 performance may result from its MAE-based pretraining, which is sensitive to domain mismatch when not fine-tuned. Object-centric models perform comparably to top baselines here, despite the simplicity of the environment.

In \textbf{LIBERO}, which features complex scenes with multiple objects, OCRs clearly outperform all other representations. VIDEOSAUR* improves over the best dense model (Theia) by +9\%, showing its ability to handle multi-object interaction.

In the \textbf{real-world} setup using LeRobot, OCRs again outperform other models. VIDEOSAUR* reaches a 70\% success rate compared to 50\% for the best dense baseline. Interestingly, the simplest model---ResNet-50 pretrained on ImageNet---also performs competitively, possibly due to its compact size and the diversity of visual pretraining data in line with the findings in \cite{dasari2023unbiasedlookdatasetsvisuomotor}. 

Comparison of all the four OCR variants enables further insights. Firstly, incorporating robotic data into pre-training proves immensely beneficial when comparing the performance of OCR models (DINOSAUR* vs DINOSAUR, VIDEOSAUR* vs VIDEOSAUR). In particular,  VIDEOSAUR* increases VIDEOSAUR's mean performance by 13, 11 and 10 points in MetaWorld, LIBERO and LeRobot environments, respectively. Secondly, putting side by side DINOSAUR* with VIDEOSAUR* which differs only by an additional modeling of the temporal dynamics while using the same pretraining data , it can be seen from Figure \ref{fig:perf}  that taking into account temporal dynamics in VIDEOSAUR* constitutes another major factor of performance gain, improving DINOSAUR* by +9 and +26 points in the LIBERO and LeRobot environments, respectively. One cannot draw similar conclusion when comparing VIDEOSAUR and DINOSAUR because their pretraining data are different. It could explain in particular the performance decrease of VIDEOSAUR over DINOSAUR in Metaworld. Specifically, robotic data provides structured scenes with fixed cameras and consistent motion patterns, such as a robotic arm moving within the frame. This contrasts with the in-the-wild dynamics used for training the original VIDEOSAUR, which often includes moving cameras and less predictable scene changes. Furthermore, the visual domain also differs substantially.

Overall, our results for Q1 confirm that object-centric models are not only effective but scalable across domains, offering performance benefits in both structured simulation tasks and noisy real-world environments.

\subsection{Q2: Do OCRs enhance generalization under visual distribution shifts?}

We now evaluate generalization to out-of-distribution conditions, including novel distractors, unseen textures, and lighting changes. Table~\ref{table:perf_gen_meta} and Table~\ref{table:perf_gen_real} report success rates for these scenarios.

In \textbf{MetaWorld}, as can be seen in Table~\ref{table:perf_gen_meta}, dense models like DINOv2 and Theia outperform ResNet on average, confirming previous findings (e.g.,\cite{burns2023makespretrainedvisualrepresentations}). However, OCRs---especially VIDEOSAUR*---consistently outperform all baselines under texture and lighting shifts. Notably, Theia performs best in the distractor scenario, likely due to CLIP-based text-image alignment helping the policy ignore irrelevant patches. Yet, Theia suffers a sharp drop in the other scenarios, where OCRs prove to be much more robust.

\begin{table}[h]
\caption{\textbf{Generalization MetaWorld.} Comparison of final results for different levels of generalization on MetaWorld. ID: performance using In-domain test data}
\label{table:perf_gen_meta}
\begin{center}
\begin{tabular}{|c|c|c|c|c|c|}
\hline
Models & Distractors  & Textures & Lighting & Overall & ID \\
\hline
ResNet-50 (pre-trained) \cite{he2015deepresiduallearningimage}  & 0.04 ± 0.02  & 0.001 ± 0.002 & 0.22 ± 0.02 & 0.10 &  \textbf{0.76} \\
R3M \cite{nair2022r3muniversalvisualrepresentation} & 0.0 ± 0.0 & 0.0 ± 0.0  & 0.0 ± 0.0 & 0.0 & 0.68 \\
DINOv2 \cite{oquab2024dinov2learningrobustvisual} &   0.11 ± 0.02  & 0.03 ± 0.01 &  0.39 ± 0.04 & 0.18 & 0.71\\
VC1 \cite{majumdar2024searchartificialvisualcortex}  & 0.06 ± 0.02  & 0.0 ± 0.0  & 0.23 ± 0.03 & 0.10 & 0.50 \\
Theia \cite{shang2024theiadistillingdiversevision} & \textbf{0.65} ± 0.10  & 0.28 ± 0.06  & 0.48 ± 0.10 & 0.47 & 0.75 \\
DINOSAUR \cite{seitzer2023bridginggaprealworldobjectcentric} & 0.21 ± 0.04 & \textbf{0.48} ± 0.06 & \textbf{0.71} ± 0.06 & 0.46 & 0.73 \\
DINOSAUR* \cite{seitzer2023bridginggaprealworldobjectcentric} & 0.46 ± 0.14 & 0.36 ± 0.05 & 0.65 ± 0.14 & 0.49 & \textbf{0.76} \\
VIDEOSAUR \cite{zadaianchuk2023objectcentriclearningrealworldvideos} & 0.29 ± 0.13  & 0.36 ± 0.10  & 0.57 ± 0.10 & 0.41 &  0.63 \\
VIDEOSAUR* \cite{zadaianchuk2023objectcentriclearningrealworldvideos} & 0.49 ± 0.13  & 0.35 ± 0.11  & 0.65 ± 0.13 & \textbf{0.50} &  \textbf{0.76} \\
\hline
\end{tabular}
\end{center}
\end{table}

In \textbf{real-world evaluations} as can be seen in Table~\ref{table:perf_gen_real}, VIDEOSAUR* shows strong robustness, with the lowest drops in the success rate under texture and lighting changes. Theia again performs best in the distractor scenario, but fails under other shifts. ResNet-50, while providing surprisingly good resilience to texture and lighting variation, lacks consistency across all shifts.

\begin{table}[h]
\caption{\textbf{Generalization real-world.} Comparison of final results for different levels of generalization on real-world. ID: performance using In-Domain test data.}
\label{table:perf_gen_real}
\begin{center}
\begin{tabular}{|c|c|c|c|c|c|}
\hline
Models & Distractors  & Textures & Lighting & Overall & ID\\
\hline
ResNet-50 (pre-trained) \cite{he2015deepresiduallearningimage}  & 0.08 & 0.22  & 0.32 & 0.21 & 0.5 \\
DINOv2 \cite{oquab2024dinov2learningrobustvisual} & 0  & 0 & 0.02 & 0.01 & 0.22 \\
Theia \cite{shang2024theiadistillingdiversevision} & \textbf{0.32} & 0.28
& 0.22 & 0.28 & 0.32\\
DINOSAUR \cite{seitzer2023bridginggaprealworldobjectcentric} & 0.18  & 0.32 & 0.30 & 0.27 & 0.40 \\
DINOSAUR* \cite{seitzer2023bridginggaprealworldobjectcentric} & 0.28  & 0.34 & 0.36 & 0.33 & 0.44 \\
VIDEOSAUR \cite{zadaianchuk2023objectcentriclearningrealworldvideos} & 0.26 & 0.44 & 0.50 & 0.40 & 0.60 \\
VIDEOSAUR* \cite{zadaianchuk2023objectcentriclearningrealworldvideos} & 0.24  & \textbf{0.5} & \textbf{0.58} & \textbf{0.44} & \textbf{0.72}\\
\hline
\end{tabular}
\end{center}
\end{table}

In summary, results for Q2 demonstrate that \textbf{object-centric representations generalize better across diverse distribution shifts}, particularly those that perturb low-level appearance. This robustness likely stems from OCRs’ ability to filter task-irrelevant background and focus on object-level structure.

\section{Conclusion}
\label{sec:conclusion}
In this work, we investigate the potential of OCRs to enhance the generalization and robustness of robotic manipulation policies. Unlike traditional visual encoders, OCR-based models introduce inductive biases that better reflect the structured nature of physical interaction, yielding consistently superior performance across a diverse set of simulated and real-world manipulation tasks.

Our results highlight the importance of rethinking visual representations to advance both the efficiency and generalization of robotic agents: moving away from flat, pixel-level features toward more structured, object-based encodings. By leveraging object-centric biases, we can bridge the gap between low-level visual input and high-level symbolic reasoning, enabling robots to better understand and interact with their surroundings.

We believe OCRs present a promising foundation for bridging the sim-to-real gap and achieving scalable, generalizable robotic control. Future research should explore how OCRs can be further integrated with diverse architectures, multimodal inputs, and self-supervised learning frameworks to maximize their scalability and downstream utility.



\clearpage

\section{Limitations}
\label{sec:limitations}
While we tried to conduct extensive and thorough experiments, several limitations remains and should be addressed in future works. 

First, the object-centric methods employed in this study do not inherently bind to specific objects and lack semantic grounding. As illustrated in Figure \ref{fig:slots}, some slots are allocated to background regions without capturing meaningful semantic content, and in some failure cases slots also capture distractors existing slots \ref{fig:slots_fail}. This limitation suggests that incorporating semantic information, e.g., affordance,  could enhance the interpretability and the practical utility of the object-centric representations. Furthermore, it may provide a pathway to improve the generalization capabilities of OCRs in the presence of distractors, as suggested by the results obtained with Theia

Second, our work does not account for the alignment with robot dynamics. We posit that integrating this aspect could significantly improve the practical applicability of our methods, particularly in light of recent advances highlighted in \cite{jiang2024robots}. Future research should explore this avenue to better align object-centric models with robotic systems.

Finally, the scope of our exploration has been constrained by the scale of the models and the datasets used for pre-training. Expanding this work to include larger models and more diverse datasets could yield more robust and generalizable results, in the light of what was done in \cite{didolkar2024zeroshotobjectcentricrepresentationlearning}. This would also allow for a more thorough evaluation of the models' capabilities and limitations.

 

\acknowledgments{}


\bibliography{example}  

\appendix

\section{Implementation details}
\label{annex:mod_detail}
\paragraph{Backbone} Following DINOSAUR \cite{seitzer2023bridginggaprealworldobjectcentric} and VIDEOSAUR \cite{zadaianchuk2023objectcentriclearningrealworldvideos} works, we use DINOv2 \cite{oquab2024dinov2learningrobustvisual} model as input backbone to our object-centric. We use the ViT-B14 version and use images of resolution 224$\times$224 as input. 

\paragraph{Object-centric layers} For DINOSAUR model, a Slot-Attention layer \cite{locatello2020objectcentriclearningslotattention} is added on top of the backbone with a classical MLP decoder, detailed in Table \ref{tab:model_detail}, which reconstruct the input patches from DINO model. For VIDEOSAUR, we follow the original paper and use a transformer \textit{predictor} which initialize slots from timestep $t$ with those of timestep $t-1$ in order to transmit dynamic information. It also uses a Slot-Attention layer on top of DINOv2 but encode here an history of $H$ frames. each frame is decoded independently with the SlotMixerDecoder introduced in the original paper \cite{zadaianchuk2023objectcentriclearningrealworldvideos}. An overview of models parameters is shown in Table \ref{tab:model_detail}.

\begin{table}[h]
\caption{Object-centric and policy model details.}
\label{tab:model_detail}
\begin{center}
\begin{tabular}{l}
\hline
DINOSAUR \cite{seitzer2023bridginggaprealworldobjectcentric} \\
\hline
\textbf{Backbone:} DINOv2 \cite{oquab2024dinov2learningrobustvisual} - ViT-B14 \\
\textbf{Object-Centric:} Slot-Attention \cite{locatello2020objectcentriclearningslotattention} - 3 iteration, Slot size=128, Num slots=10 \\
\textbf{Decoder:} MLP Decoder - 3 layers, Hidden dim=768 \\
\hline
VIDEOSAUR \cite{zadaianchuk2023objectcentriclearningrealworldvideos} \\
\hline
\textbf{Backbone:} DINOv2 \cite{oquab2024dinov2learningrobustvisual} - ViT-B14 \\
\textbf{Object-Centric:} Slot-Attention \cite{locatello2020objectcentriclearningslotattention} - 3 iteration, Slot size=128, Num slots=10 \\
\textbf{Prediction layer:} Transformer(Num layers=1, Hidden dim=512, Num heads=4) \\
\textbf{Decoder:} Slot Mixer Decoder \cite{zadaianchuk2023objectcentriclearningrealworldvideos} - Allocator=Transformer(Num layers=2, Hidden dim=512, Num heads=4), \\
Predictor=Transformer(Num layers=1, Hidden dim=512, Num heads=4) \\
\hline
\\
\hline
BAKU \cite{haldar2024bakuefficienttransformermultitask} \\
\hline
\textbf{Observation trunk:} Transformer(Num layers=8, Num heads=4) (minGPT) \\
\textbf{Action Head:} MLP(Hidden dim=256) (deterministic)\\
Action chunking True (10 chunks) \\
History len=4 \\
\hline
ACT \cite{cadene2024lerobot}\\
\hline
\textbf{Task embedding:} ModernBERT (Base version) \cite{warner2024smarterbetterfasterlonger} \\
\textbf{VAE Encoder:} VAE(Latent dim=32, Num layers=4)\\
\textbf{Transformer Encoder:} Transformer Encoder(Num layers=4, Num Heads=8, Hidden dim=512)\\
\textbf{Transformer Decoder:} Transformer Decoder(Num layers=1, Num Heads=8, Hidden dim=512)\\
Action chunking True (100 chunks) \\
Temporal Ensembling False \\
\hline
\end{tabular}
\end{center}
\end{table}

\paragraph{Policy models}
In simulation, we adapt the BAKU \cite{haldar2024bakuefficienttransformermultitask} model. We specifically use the model combining a Transformer Observation trunk with a MLP deterministic action head, as it was their best performing combination. We modified the observation trunk in order to accept different input shapes instead of the original single feature. Our version can either have one single vector per view (global), several vectors per view (object-centric) or a dense set of features (dense). These vectors are interleaved one view after the other, followed by proprioception and combined with other timesteps for an history len of $H$. The task embedding is prepended at the beginning of the sequence. 

For real-world experiments, we modified the ACT \cite{zhao2023learningfinegrainedbimanualmanipulation} implementation of the LeRobot \cite{cadene2024lerobot} library. First, as the model is made to only handle single-task, we compute a task embedding with ModernBERT \cite{warner2024smarterbetterfasterlonger} and prepend the embedding vector at the input of the transformer decoder before the observation features. Then, the original model was made to take a dense feature representation from any single view as input. We made some changes in order to either accept a single global vector or several slots. This required mainly to change the 2D positional embedding of the features to a 1D positional embedding but also to modify the projection layer.

\section{Training}
\label{annex:training}
The pretraining of our DINOSAUR and VIDEOSAUR models on robotic data is performed on a single NVIDIA H100 GPU. All configurations are exposed on Table \ref{tab:train_detail}.

\begin{table}[h]
\caption{DINOSAUR and VIDEOSAUR pre-training configurations.}
\label{tab:train_detail}
\begin{center}
\begin{tabular}{l c}
\hline
Hyperparameters & \\
\hline
$\#$ GPUs & 1 \\
Batch size & 128 \\
Learning rate & 1e-4\\
LR schedule & Linear \\
Weight decay & 0.0 \\
Optimizer & AdamW\\
Betas & (0.9, 0.999) \\
Training steps & 100000\\
Warmup steps & 2500 \\
Warmup LR Schedule & 100000 \\
Gradient cliping & $\emptyset$\\
Total GPU hours & 80 \\
\hline
\end{tabular}
\end{center}
\end{table}

\section{Pre-training data ablation}
\label{annex:data}
Each of the visual models selected is pre-trained on different datasets, ranging from ImageNet \cite{russakovsky2015imagenetlargescalevisual}, to egocentric manipulation data \cite{grauman2022ego4dworld3000hours}. The object-centric models were originally trained on \textit{in-the-wild} images and videos datasets. In our work, we decided to evaluate whether training them on different kind of data might help to improve their downstream performances for robotic manipulation. Thus, we retrained from scratch these models on the large-scale robotic datasets BridgeV2 \cite{walke2024bridgedatav2datasetrobot}, Fractal \cite{brohan2023rt1roboticstransformerrealworld} and DROID \cite{khazatsky2024droidlargescaleinthewildrobot} but also on a Mixture combining all of these data, in order to better align their representations with the task dynamics we aim to solve.

\begin{table}[h]
\caption{\textbf{Pre-training data evaluation.} Comparison of the pre-training data on the final global performance for VIDEOSAUR \cite{zadaianchuk2023objectcentriclearningrealworldvideos} model.}
\label{table:abl_data}
\begin{center}
\begin{tabular}{|c||c|c|c|}
\hline
Pre-training data  & Metaworld & LIBERO  & LeRobot \\
\hline
Youtube-VOS \cite{xu2018youtubevoslargescalevideoobject}  & 0.63  & 0.77 & 0.58 \\
\hline
BridgeV2 (B) \cite{walke2024bridgedatav2datasetrobot} & 0.77 & 0.84 & 0.54 \\
\hline
Fractal (F) \cite{brohan2023rt1roboticstransformerrealworld} & 0.71 & 0.60 & 0.42 \\
\hline
DROID (D) \cite{khazatsky2024droidlargescaleinthewildrobot} & \textbf{0.79} & 0.78 & 0.28 \\
\hline
(Ours) Mixture (D+B+F) & 0.76 & \textbf{0.86} & \textbf{0.7} \\
\hline
\end{tabular}
\end{center}
\end{table}

\section{Environments details}
\label{annex:env_detail}

This section details the two simulation environments - MetaWorld \cite{yu2021metaworldbenchmarkevaluationmultitask} and LIBERO \cite{liu2023libero} - and the real-world setup - using the LeRobot library \cite{cadene2024lerobot} - we use to evaluate our different visual models.

\textbf{MetaWorld}  \cite{yu2021metaworldbenchmarkevaluationmultitask} is a benchmark built on top of the MuJoCo engine \cite{todorov2012mujoco}, featuring a collection of tabletop manipulation tasks using a Sawyer arm. Following \cite{jiang2024robots}, we selected a subset of 10 tasks and collected 50 demonstrations per task using the official provided policies. MetaWorld is a widely used robotic benchmark in the community, and permits to quickly evaluate and compare our models in different scenarios with existing baselines, even with different generalization levels thanks to the setup proposed in \cite{xie2023decomposinggeneralizationgapimitation}. 
The set of 10 selected tasks are the following:
\begin{itemize}
    \item \texttt{Assembly} ($\mathcal{A} \in \mathbb{R}^4$): the goal is to grasp a nut and place it around a stick.
    \item \texttt{Bin Picking} ($\mathcal{A} \in \mathbb{R}^4$): the goal is to pick a green cube from a blue bin and to place it into a red bin.
    \item \texttt{Button Press} ($\mathcal{A} \in \mathbb{R}^4$): the goald it to press the button.
    \item \texttt{Disassemble} ($\mathcal{A} \in \mathbb{R}^4$): the goal is remove a nut from around a stick.
    \item \texttt{Drawer Open} ($\mathcal{A} \in \mathbb{R}^4$): the goal is to open the drawer.
    \item \texttt{Hammer} ($\mathcal{A} \in \mathbb{R}^4$): the goal is to drive a screw into the wall using a hammer.
    \item \texttt{Pick Place Wall} ($\mathcal{A} \in \mathbb{R}^4$): the goal is to pick a cube and place it at a certain position situated behind a wall. 
    \item \texttt{Shelf Place} ($\mathcal{A} \in \mathbb{R}^4$): the goal is to pick a deformable blue puck and to place it into a wooden shelf.
    \item \texttt{Stick Pull} ($\mathcal{A} \in \mathbb{R}^4$): the goal is to pull a pitcher using a stick.
    \item \texttt{Stick Push} ($\mathcal{A} \in \mathbb{R}^4$): the goal is to push a pitcher using a stick.
\end{itemize}

\textbf{LIBERO} \cite{liu2023libero} is a benchmark built on top of Robomimic \cite{robomimic2021},  comprising multiple task suites for diverse purposes. We selected the LIBERO-90 suite, which includes 90 tasks, each with 50 demonstrations. The suite spans three distinct environments: kitchen, office, and living room; offering a broad range of objects and task variations. This benchmark is particularly suitable to evaluate our object-centric models as it implies multi-object scenes as shown in Figure \ref{fig:tasks} in more visually diverse environments than MetaWorld and permits to better illustrate learning in complex scenarios. 

\begin{figure}[h]
    \centering
    \includegraphics[width=\textwidth]{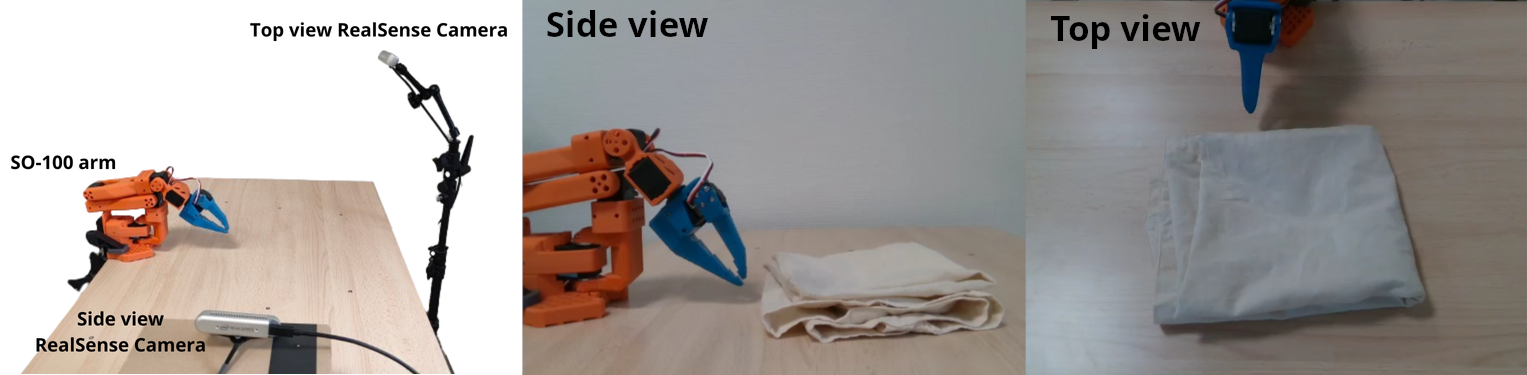}
    \caption{\textbf{Real world setup.} Our setup is based on the LeRobot library. We use a SO-100 arm on a tabletop environment with two realsense cameras, one with a top overview of the scene and a second with a side view.}
    \label{fig:setup}
\end{figure}

\textbf{LeRobot} \cite{cadene2024lerobot} is an open-source library providing models and tools to build a low-cost robotic arm with a teleoperation leader ($\sim$300\$) - SO-100- enabling easy real-world experiment replication. We selected a set of five tasks and collected 30 demonstrations per task using the setup shown in Figure \ref{fig:setup}.

\begin{figure}[h]
    \centering
    \includegraphics[width=\textwidth]{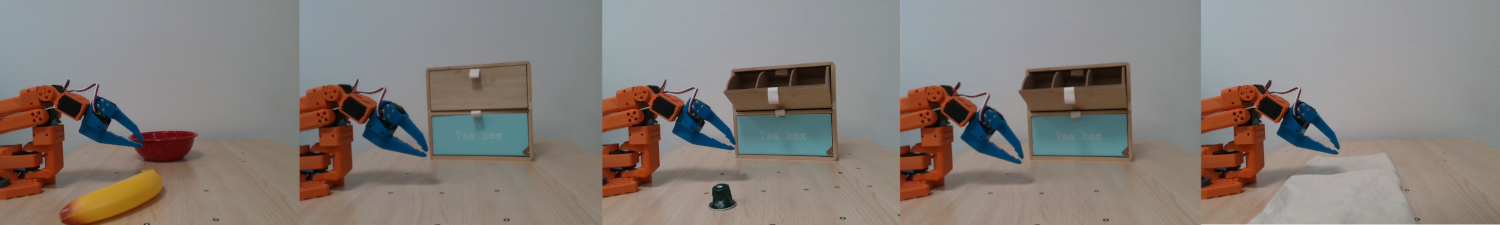}
    \caption{\textbf{Overview tasks real-world.} From left to right: \texttt{Banana bowl}, \texttt{Open drawer}, \texttt{Pick coffee}, \texttt{Close drawer}, \texttt{Fold bag}.}
    \label{fig:tasks_real}
\end{figure}

For our real world environments, we selected a set of 5 tasks designed to evaluate a range of essential robotic skills:
\begin{itemize}
    \item \texttt{Open Drawer} ($\mathcal{A} \in \mathbb{R}^7$): Open the top drawer of the teabox
    \item \texttt{Close Drawer} ($\mathcal{A} \in \mathbb{R}^7$): Close the top drawer of the teabox
    \item \texttt{Pick Coffee} ($\mathcal{A} \in \mathbb{R}^7$): Pick the coffe cap and place it into the top drawer of the teabox
    \item \texttt{Fold bag} ($\mathcal{A} \in \mathbb{R}^7$): Fold the bag from bottom-right corner
    \item \texttt{Banana bowl} ($\mathcal{A} \in \mathbb{R}^7$): Put the banana into the red bowl
\end{itemize}
A visualization of all the tasks is performed on Figure \ref{fig:tasks_real}.
Each task necessitates distinct manipulation skills, including picking, placing, opening, closing and folding. These skills involve interacting with various type of objects: rigid, deformable and articulated objects.
These tasks are representative of common household activities, and successful execution indicates a robot's potential to assist in daily living tasks. This comprehensive evaluation ensures that the robotic model can handle the complexity and variability of real-world environments.

\begin{figure}[h]
    \centering
    \includegraphics[width=\textwidth]{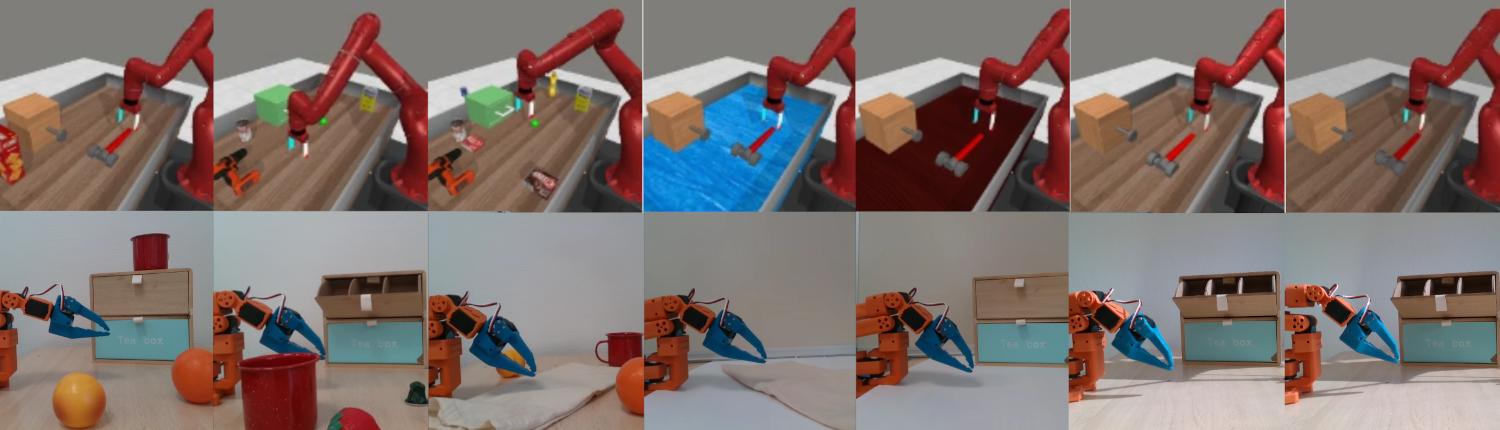}
    \caption{\textbf{Overview of different generalization levels.} From left to right: new distractors, new textures of the table, new  lighting conditions. Top row: Metaworld. Bottom row: Real-World}
    \label{fig:gen}
\end{figure}

\section{Baselines details}
\label{annex:baseline_detail}
This section details the 7 state-of-the-art visual models we compare in our experiments. It covers different architecture types (Convolutionnal Neural Networks and Vision Transformers) but also different data (natural images, manipulation data, in-the-wild videos). 

(1) \textbf{ResNet-50} \cite{he2015deepresiduallearningimage}: A classical baseline model pre-trained on ImageNet, widely used for its robust performance in various computer vision tasks. It serves as a standard for comparison due to its simplicity and effectiveness. \newline
(2) \textbf{R3M} \cite{nair2022r3muniversalvisualrepresentation}: This model, based on ResNet-50 architecture, leverages egocentric manipulation data and time-contrastive learning, making it particularly effective for robotic tasks that require understanding temporal sequences and object interactions. \newline
(3) \textbf{DINOv2} \cite{oquab2024dinov2learningrobustvisual}: Utilizes a Vision Transformer architecture with self-supervised learning objectives, achieving state-of-the-art performance across numerous downstream tasks. Its training on a large-scale curated dataset enhances its robustness and generalization capabilities. In our experiments, we tried the model with Global and Dense feature representation as input to the policy. As the Global representation was always outperforming the other alternative we only show in the experimental sections the result of the Global representation model.  \newline
(4) \textbf{VC-1} \cite{majumdar2024searchartificialvisualcortex}: Combines the strengths of Vision Transformers with a Masked-Auto Encoder objective, pre-trained on both manipulation data and ImageNet. This model is well-suited for tasks requiring fine-grained visual understanding and manipulation. \newline
(5) \textbf{Theia} \cite{shang2024theiadistillingdiversevision}: Distilled from multiple state-of-the-art models, Theia integrates various modalities such as depth, segmentation, and language. Its diverse training objectives make it highly versatile and effective in complex robotic benchmarks. In our experiments, we took the best performing Theia model : CDiV, the model distilled from CLIP \cite{radford2021learningtransferablevisualmodels}, DINOv2 \cite{oquab2024dinov2learningrobustvisual} and a pre-trained ViT on ImageNet.  \newline
(6) \textbf{DINOSAUR} \cite{seitzer2023bridginggaprealworldobjectcentric}: Built on top of DINOv2, this model focuses on object-centric representations using feature reconstruction. It adapts Slot-Attention to real-world scenarios, enhancing its ability to handle object-specific tasks. \newline
(7) \textbf{VIDEOSAUR} \cite{zadaianchuk2023objectcentriclearningrealworldvideos}: An extension of DINOSAUR to video data, incorporating a temporal alignment loss. This model excels in unsupervised image segmentation in real-world settings, leveraging temporal information for improved performance.

\section{Per task evaluation}
\label{annex:per_task}
To better understand the success and failure of the different models, we also analyze task-level success rates, as shown in Table \ref{table:metaworld_task} for MetaWorld and Table \ref{table:real_task} for real-world experiments.

\begin{table}[h]
\caption{\textbf{Per task Metaworld.} Comparison of the results per task on the MetaWorld environment. A: Assembly; BP: Bin Picking; BuP: Button Press; D: Dissamble; DO: Drawer Open; H: Hammer, P\&P: Pick and Place; SP: Shelf place; StP:Stick Pull; StPus: Stick Push}
\label{table:metaworld_task}
\begin{center}
\begin{tabular}{|c|c|c|c|c|c|}
\hline
Models & A  & BP & BuP & D & DO  \\
\hline
ResNet-50 (pre-trained) \cite{he2015deepresiduallearningimage}  & 0.73 & 0.37  & 1.0 & 0.62 & 1.0 \\
R3M \cite{nair2022r3muniversalvisualrepresentation} & 0.38 & 0.31 & 1.0 & 0.35 & 1.0\\
DINOv2 \cite{oquab2024dinov2learningrobustvisual} & 0.57 & 0.41 & 1.0 & 0.54 & 1.0 \\
VC1 \cite{majumdar2024searchartificialvisualcortex}  & 0.11  & 0.15  & 1.0 & 0.47 & 1.0\\
Theia \cite{shang2024theiadistillingdiversevision} & 0.36  & 0.52  & 1.0 & \textbf{0.72} & 1.0 \\
DINOSAUR \cite{seitzer2023bridginggaprealworldobjectcentric} & \textbf{0.90} & \textbf{0.53} & 1.0 & 0.60 & 1.0 \\
DINOSAUR* \cite{seitzer2023bridginggaprealworldobjectcentric} & 0.80 & 0.30 & 1.0 & 1.0 & 1.0 \\
VIDEOSAUR \cite{zadaianchuk2023objectcentriclearningrealworldvideos} & 0.20 & 0.50 & 1.0 & 0.67 & 1.0 \\
VIDEOSAUR* \cite{zadaianchuk2023objectcentriclearningrealworldvideos} & 0.80 & 0.47 & 1.0 & 0.60 &  1.0 \\
\hline
& H  & P\&P & SP & StP & StPus  \\
\hline
ResNet-50 (pre-trained) \cite{he2015deepresiduallearningimage}  & 0.99  & 0.71  & 0.39 & 0.61 & 1.0 \\
R3M \cite{nair2022r3muniversalvisualrepresentation} & 1.0 & 0.39 & 0.29 & 0.70 & 1.0 \\
DINOv2 \cite{oquab2024dinov2learningrobustvisual} & 0.97 & 0.74 & 0.09 & 0.63 & 1.0 \\
VC1 \cite{majumdar2024searchartificialvisualcortex}  & 0.85  & 0.49  & 0.07 & 0.49 & 1.0 \\
Theia \cite{shang2024theiadistillingdiversevision} & 1.0  & \textbf{0.75} & \textbf{0.55} & 0.57 & 1.0  \\
DINOSAUR \cite{seitzer2023bridginggaprealworldobjectcentric} & 1.0 & 0.40 & 0.20 & 0.80 & 1.0 \\
DINOSAUR* \cite{seitzer2023bridginggaprealworldobjectcentric} & 0.80 & 0.60 & 0.20 & 0.40 & 1.0 \\
VIDEOSAUR \cite{zadaianchuk2023objectcentriclearningrealworldvideos} & 0.60 & 0.20 & 0.40 & 0.80 & 1.0 \\
VIDEOSAUR* \cite{zadaianchuk2023objectcentriclearningrealworldvideos} & 1.0 & 0.60 & 0.27 & \textbf{0.90} & 1.0\\
\hline
\end{tabular}
\end{center}
\end{table}

As can be seen from Table \ref{table:metaworld_task}, every models primarly fails on three tasks: assembly, bin picking and shelf place, each requiring precise object manipulation. The object-centric model performs on par with the best model, Theia \cite{shang2024theiadistillingdiversevision}, in most scenarios but underperform in the shelf place task. We make the assumption that this performance gap is due to the fact that the cube to grasp in this environment is of a small size, which can be hard for our current object-centric model to segment. This is left to be solved with future works. However, in the assembly task, that implies a bigger object, the object-centric model are outperforming the other models.

\begin{table}[h]
\caption{\textbf{Per task real-world.} Comparison of the results per task on the real-world environment. OD: Open Drawer; CD: Close Drawer; PCD: Put Coffee into Drawer; BB: Banana to Bowl, F: Fold}
\label{table:real_task}
\begin{center}
\begin{tabular}{|c|c|c|c|c|c|c|}
\hline
Models & OD  & CD & PCD & BB & F & Overall \\
\hline
ResNet-50 (pre-trained) \cite{he2015deepresiduallearningimage}  & 0/10  & \textbf{10/10}  & \textbf{4/10} & \textbf{7/10} & 4/10 & 25/50 \\
R3M \cite{nair2022r3muniversalvisualrepresentation} & 5/10 & 1/10 & 1/10 & 0/10 & 0/10 & 7/50 \\
DINOv2 \cite{oquab2024dinov2learningrobustvisual} & 9/10 & 0/10 & 0/10 & 0/10 & 2/10 & 11/50 \\
VC1 \cite{majumdar2024searchartificialvisualcortex}  & 9/10  & 1/10  & 0/10 & 0/10 & 1/10 & 11/50 \\
Theia \cite{shang2024theiadistillingdiversevision} & \textbf{10/10}  & 5/10  & 0/10 & 0/10 & 1/10 & 16/50  \\
DINOSAUR \cite{seitzer2023bridginggaprealworldobjectcentric} & 5/10 & 4/10 & 1/10 & 6/10 & 4/10 & 20/50 \\
DINOSAUR* \cite{seitzer2023bridginggaprealworldobjectcentric} & \textbf{10/10}  & 6/10  & 0/10 & 4/10 & 2/10 & 22/50  \\
VIDEOSAUR \cite{seitzer2023bridginggaprealworldobjectcentric} & \textbf{10/10}  & 9/10  & 1/10 & 6/10 & 4/10 & 30/50  \\
VIDEOSAUR* \cite{zadaianchuk2023objectcentriclearningrealworldvideos} & \textbf{10/10} & \textbf{10/10} & \textbf{4/10} & \textbf{7/10} &  \textbf{5/10} & \textbf{36/50} \\
\hline
\end{tabular}
\end{center}
\end{table}

In the real-world scenario, the object-centric model outperforms all other models in all tasks as shown in Table \ref{table:real_task}.

\section{Visualization of slots}
\begin{figure}[h]
    \centering
    \includegraphics[width=\textwidth]{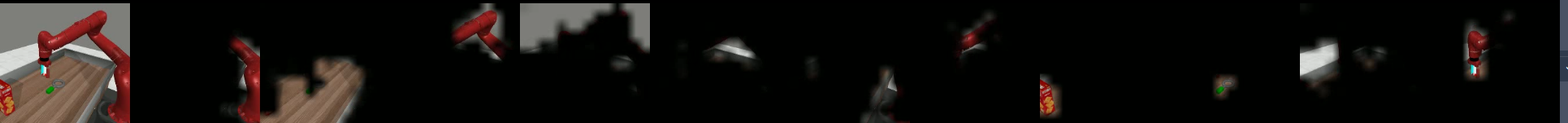}
    \caption{\textbf{Slots visualization.} Visualization of a set of slots (12 slots) extracted from VIDEOSAUR* model on the easy distractor setup in Metaworld}
    \label{fig:slots}
\end{figure}

We visualize several slot decomposition of our VIDEOSAUR* model on different tasks on Figure \ref{fig:slots}. Note that the model has never seen the provided data before as it has been pre-trained on frozen to learn subsequent policy. 

\section{Failure cases visualization}
\begin{figure}[h]
    \centering
    \includegraphics[width=\textwidth]{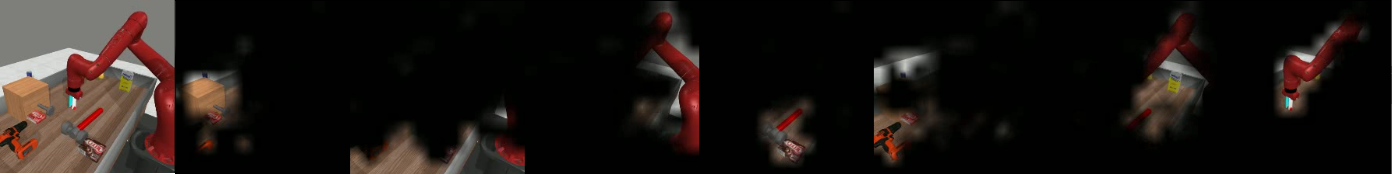}
    \caption{\textbf{Failure slots visualization.} Visualization of a set of slots (8 slots) extracted from VIDEOSAUR* model on the hard distractor setup in Metaworld. The slot that should handle the hammer to perform the task also capture the brown box, leading to noise in the subsequent slot representation.}
    \label{fig:slots_fail}
\end{figure}

In order to better understand why the object-centric model fails in the distractor scenarios, we decided to check in detail the slots as shown on Figure \ref{fig:slots_fail}. 
\end{document}